\documentclass[sigconf]{acmart}

\usepackage{amsfonts}
\usepackage{multirow}
\usepackage{subfigure}
\usepackage{graphicx}
\usepackage{caption}
\usepackage{amsmath}

\AtBeginDocument{%
  \providecommand\BibTeX{{%
    \normalfont B\kern-0.5em{\scshape i\kern-0.25em b}\kern-0.8em\TeX}}}

\begin{document}

\title[TransModality: An End2End Fusion Method with Transformer]{TransModality: An End2End Fusion Method with Transformer for Multimodal Sentiment Analysis}


\author{Zilong Wang, Zhaohong Wan, and Xiaojun Wan}
\affiliation{%
  \institution{Wangxuan Institute of Computer Technology, Peking University}
  \institution{The MOE Key Laboratory of Computational Linguistics, Peking University}}
\email{{wang_zilong, xmwzh, wanxiaojun}@pku.edu.cn}




\renewcommand{\shortauthors}{Zilong Wang, Zhaohong Wan, Xiaojun Wan}

\begin{abstract}
  Multimodal sentiment analysis is an important research area that predicts speaker's sentiment tendency through features extracted from textual, visual and acoustic modalities. The central challenge is the fusion method of the multimodal information. A variety of fusion methods have been proposed, but few of them adopt end-to-end translation models to mine the subtle correlation between modalities. Enlightened by recent success of Transformer in the area of machine translation, we propose a new fusion method, TransModality, to address the task of multimodal sentiment analysis. We assume that translation between modalities contributes to a better joint representation of speaker's utterance. With Transformer, the learned features embody the information both from the source modality and the target modality. We validate our model on multiple multimodal datasets: CMU-MOSI, MELD, IEMOCAP. The experiments show that our proposed method achieves the state-of-the-art performance.
\end{abstract}





\maketitle

\section{Introduction}
	Semantic analysis has been a hot research topic for many years. Most traditional methods are based on texts \cite{hu2004mining,liu2012sentiment,mohammad2013nrc}, since textural materials are easy to get. Efforts have been made to obtain materials from other modalities, such as collecting video scenes from daily TV series \cite{poria2018meld} and asking the professionals to perform improvisations or scripted scenarios \cite{busso2008iemocap}. Besides, emerging social media, such as YouTube, offer a new resource of multimodal materials, and the new multimodal materials are closer to real human.
	
	Therefore, multiple datasets are available, not only from the actor's performance \cite{busso2008iemocap, poria2018meld}, but also from the social media videos \cite{zadeh2016mosi}. Thanks to the datasets, multimodal sentiment analysis has attracted more and more attention these years.

	All of these resources, TV series, actors' performance and social media videos, consist of information from not only the textual modality but the acoustic and visual modality as well. Intuitively, information from acoustic or visual modalities will certainly contribute to a better prediction in sentiment analysis because it can better deal with language ambiguity. It is a non-trivial task to distinguish ambiguity only from textual information. If visual or acoustic information is considered, it can be much easier. An example from CMU-MOSI, one of datasets used, is provided below \footnote{This example is from Clx4VXItLTE video segment.}. The speaker in the video is talking about an interesting horror movie and giving her opinion on the movie. The utterance is ambiguous and will probably be predicted as \textit{negative} only through the textual information. But given smiling face and happy tone, we know the utterance is \textit{positive} and describes something interesting or surprising but not frightening in the movie plot.	With multimodal information, speaker's sentiment and emotion can be analyzed more precisely and properly.
	
	\begin{figure}[h]
		\centering
		\includegraphics[width=0.8\linewidth]{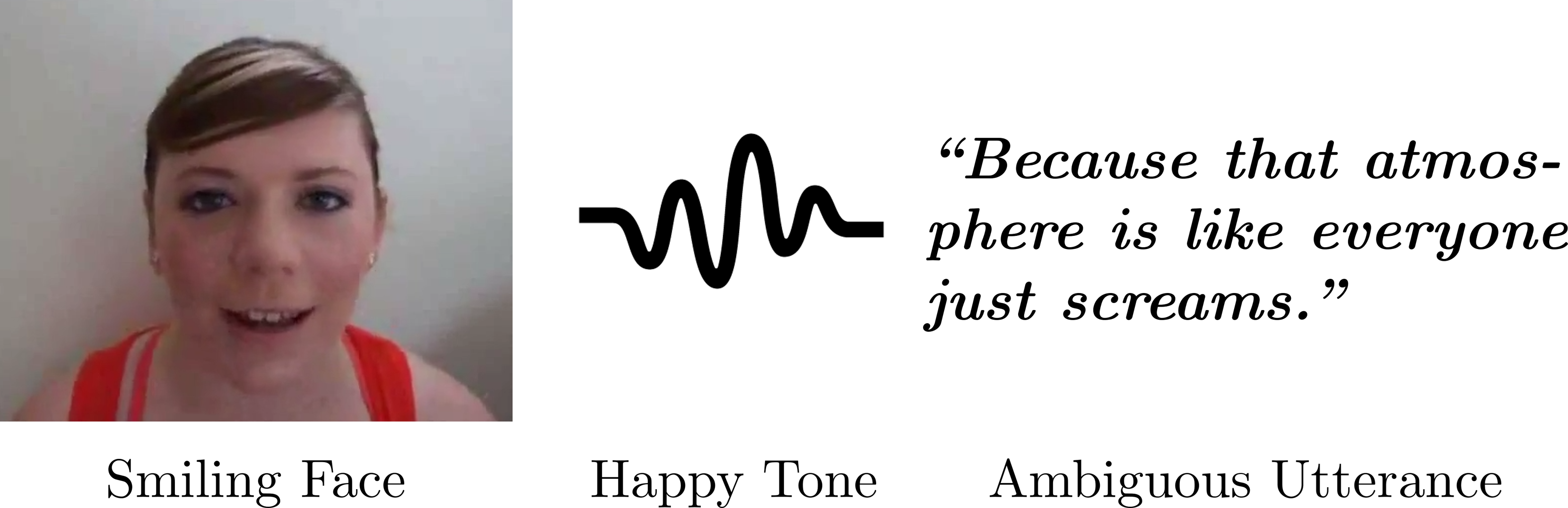}
		\caption{An example from CMU-MOSI dataset}
		\label{fig:part}
	\end{figure}

	Although modalities are useful, their distinct properties make it non-trivial to fully utilize them. It must be noted that not all modalities play equal roles in sentiment analysis. Since sentiment analysis with textual features has long been studied, textual features serve as key features for prediction. By contrast, visual and acoustic features cannot predict sentiment tendency very well. So, to prevent the interference from the visual and acoustic features, they are used as auxiliary features to improve textual sentiment analysis \cite{chen2017multimodal}. How to blend features of different modalities and improve the performance is a major challenge for the multimodal sentiment analysis task.
	
	In this paper, we propose a novel method, TransModality, to fuse multimodal features with end-to-end translation models for multimodal sentiment analysis. We select Transformer \cite{vaswani2017attention} for translation. Transformer uses attention mechanism to model the relation between source and target language. In our proposed method, features of different modalities are treated as source and target of Transformer models, respectively. We hypothesize that translation between modalities helps modality fusion. The feature of one modality is encoded first by Transformer. And the feature of another modality is decoded from the encoded feature as result. So the encoded feature embodies information from both source modality and target modality. To improve the performance of translation and robustness of our model, we adopt the parallel translation, which means we fuse textual features with acoustic features and fuse textual features with visual features independently. We also train our model with Forward Translation and Backward Translation, which means translating one modality to another and backward. A joint loss considering both classification and translation is used to train our model.
	
	We conduct experiments on three multimodal sentiment analysis datasets - CMU-MOSI \cite{zadeh2016mosi}, MELD \cite{poria2018meld} and IEMOCAP \cite{busso2008iemocap}. These datasets are widely used as benchmark of multimodal sentiment analysis. Experimental results show the efficacy of our proposed method TransModality which outperforms several strong baseline models. We also analyze the learning behavior of translation models in our method.
	
	Our contributions are summarized as follows:
	\begin{itemize}
		\item We propose a novel method TransModality to address the challenging task of multimodal sentiment analysis, by translation between modalities with end-to-end translation model - Transformer.
		\item We adopt Forward and Backward Translation, which conduct translation from one modality to another and backward, to better fuse multimodal features and we prove the effectiveness through experiments.
		\item Our proposed method achieves the state-of-the-art performance on three multimodal datasets: CMU-MOSI, MELD, IEMOCAP.
	\end{itemize}
	
	\section{Related Work}
	Sentiment analysis has been studied as the basic task in the area of natural language processing. Most approaches focused on textual materials and used CNN \cite{kim2014convolutional,krizhevsky2012imagenet} or RNN \cite{hochreiter1997long,chung2014empirical,liu2016recurrent,abdul2017emonet} to deal with the task. These approaches achieve some progress in this area and provide experience and inspiration for later research.

	Over the last few years, multimodal sentiment analysis gained many interests. Researchers focus on how to utilize multimodal features to improve text-based approaches \cite{lian2018investigation}.
	
	Many fusion methods have been proposed. \citet{poria2017context} simply feeds the concatenation of unimodal features into an LSTM. This early work demonstrates the out-performance of multimodal features and pushes some progress towards this direction. Some advanced approaches are proposed to deal with the fusion challenge as well. \citet{zadeh2017tensor} explicitly models the unimodal, bimodal and trimodal interactions using a 3-fold Cartesian product from modality embeddings to create a new joint feature. The new feature is considered as better organized representation of the original features. \citet{liu2018efficient} proposes an efficient low-rank weight decomposition method to obtain joint representation in reduced computational complexity.
	
	Recently, various neural network fusion methods have been proposed. \citet{ghosal2018contextual} uses RNN to extract contextual information and applies attention mechanism on multimodal features to obtain better utterance representation. \citet{chen2017multimodal} proposes a key idea that some multimodal features may be redundant and interfere with other features. So they adopt reinforcement learning to train a gate to filter out noisy modalities and refine the joint representation. \citet{pham2018seq2seq2sentiment,pham2018found} extends the usage of seq2seq model to the realm of multimodal learning. They assume that the intermediate representation of this model is close to the joint representation.
	
	Most works mentioned above conducts simple or direct mathematical calculation between unimodal features. The calculation may confuse the unimodal features and harm sentiment prediction. By contrast, \citet{pham2018seq2seq2sentiment} and \citet{pham2018found} solve the problem through indirect fusion method with seq2seq models. Enlightened by previous work, we follow the idea and propose a new end-to-end fusion method with Transformer for multimodal sentiment analysis. The main difference between our method and the existing work is that our framework not only abandons the simple and direct fusion methods, embraces the new idea of indirect fusion, but also leverages the state-of-the-art translation model Transformer and discovers its potential in multi-modality fusion. Compared with the sequential translation method in \citet{pham2018seq2seq2sentiment,pham2018found}, i.e. from textual to acoustic and then to visual modality, our model conducts the translation in parallel way. We design two independent modality fusion cells to blend textual features with acoustic features, and blend textual features with visual features, respectively.  This will help to eliminate the interference between different modalities. We also add extra components to the architecture, including contextual information extraction and joint feature concatenation, and propose the Forward and Backward Translation, i.e. translating features of one modality to another and backward, to better fuse the multimodal information. To the best of our knowledge, our current work is the very first of its kind that focus on modality fusion through translation with Transformer.
	
	\section{Proposed Model}
	In our proposed model, we aim to combine multimodal features for better sentiment prediction. We believe that the sentiment tendency depends not only on the features from each modality itself, but also on the interrelated relationship between them. So we utilize Transformer to translate between different modalities and learn the joint representation for utterances. In this section, we will introduce the architecture of our model.
	
	\subsection{Problem Definition and Notation}
	A multimodal dataset consists of several videos, and each video is separated into utterance-level segments. So a video $\mathbf{V}$ is denoted as $\mathbf{V}=(\mathbf{X}_1, \mathbf{X}_2, ..., \mathbf{X}_N)$ where $N$ is the number of utterances in the video, and $\mathbf{X}_i$ is one of the utterance-level segments. For each utterance $\mathbf{X}_i (1\le i \le N)$, it has features of multiple modalities, defined as $\mathbf{X}_i = (\mathbf{X}_i^t, \mathbf{X}_i^v, \mathbf{X}_i^a)$ for the textual, visual, acoustic modalities respectively. The features of different modalities are utterance-level feature vectors extracted from the textual, visual, acoustic modalities respectively, which will be described in detail for each dataset in Section 4.1. The dimensions of features are denoted as $d^{t}$, $d^{v}$, $d^{a}$, respectively. The corresponding sentiment label for this utterance is denoted as $y_i$.
	
	So our task is to build a model to predict the $\mathbf{y} = (y_1, y_2, ..., y_N)$ through the multimodal features of a video $\mathbf{V}=(\mathbf{X}_1, \mathbf{X}_2, ..., \mathbf{X}_N)$.
	\begin{figure}[t]
		\centering
		\includegraphics[width=0.5\linewidth]{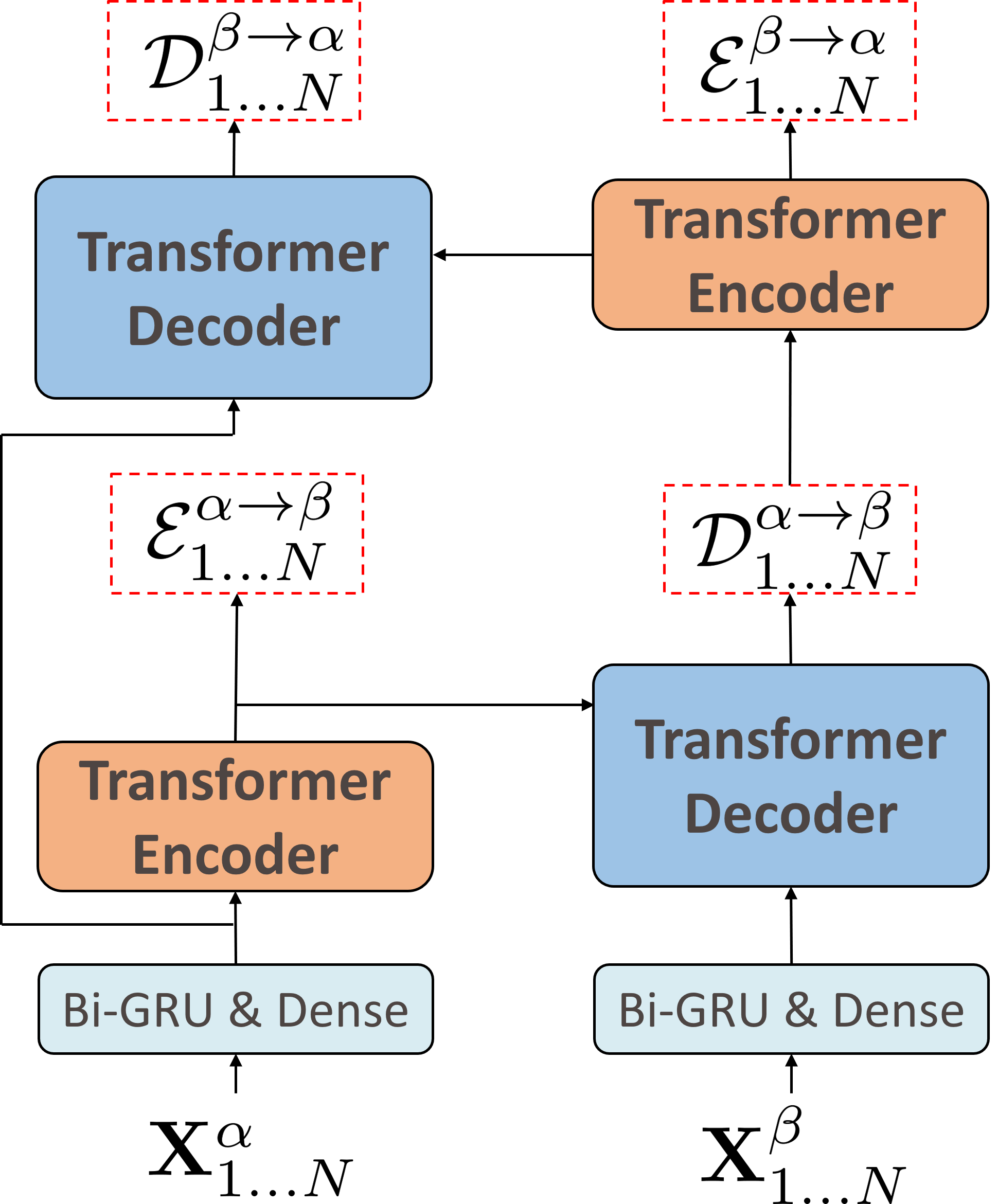}
		\caption{Modality Fusion Cell: $\mathbf{X}^{\alpha}_{1}...\mathbf{X}^{\alpha}_{N}$ and $\mathbf{X}^{\beta}_{1}...\mathbf{X}^{\beta}_{N}$ are the features of modalities involved.}
		\label{fig:mfc}
	\end{figure}
	
	\subsection{Modality Fusion Cell}
	Modality Fusion Cell is the key component of our model to perform modality fusion. It leverages the translation model - Transformer and learns the joint features of two modalities involved.
	
	We denote the two modalities involved as $\alpha$ and $\beta$, where $\alpha, \beta \in \{t,v,a\}$, and the features of these two modalities for the $N$ utterances in a video as $(\mathbf{X}_1^{\alpha}, \mathbf{X}_2^{\alpha},...,\mathbf{X}_N^{\alpha})$ and $(\mathbf{X}_1^{\beta},\mathbf{X}_2^{\beta},...,\mathbf{X}_N^{\beta})$.
	
	To enhance the fusion performance of our method, a \textit{Forward} and a \textit{Backward} Translation are conducted, which means we first translate modality $\alpha$ into modality $\beta$ (\textit{forward}) and then translate modality $\beta$ into modality $\alpha$ (\textit{backward}). So two Transformer models are used in one Modality Fusion Cell. We also use bi-directional Gated Recurrent Units (GRU) to extract contextual features before translation step.
	
	\paragraph{Contextual Information Extraction}
	We conduct bi-directional GRUs and fully connected dense layers on features of modalities $\alpha$ and $\beta$, separately. In this way, we extract the contextual information in the hidden states. Then we project the hidden states into a hyperspace of fixed dimension.
	\begin{equation}
		\mathbf{H}^{\lambda}_i = \mathrm{BiGRU}^{\lambda}(\mathbf{X}^{\lambda}_{i})
	\end{equation}
	\begin{equation}
		\mathbf{D}^{\lambda}_i = \tanh(W^{\lambda}\mathbf{H}^{\lambda}_i + b^{\lambda})
	\end{equation}
	where $\lambda \in \{\alpha, \beta\}$, $W^{\lambda}$ is the weight matrix, $b^{\lambda}$ is the bias, and $\mathbf{H}^{\lambda}_i$ is the hidden state. 
	
	\paragraph{Forward Translation}
	Forward Translation aims to blend modality $\beta$ with modality $\alpha$ by translating $(\mathbf{D}^{\alpha}_1...\mathbf{D}^{\alpha}_N)$ into $(\mathbf{X}^{\beta}_1...\mathbf{X}^{\beta}_N)$ with Forward Transformer. Transformer Encoder and Decoder are used here. The encoder's output is denoted as $(\mathcal{E}^{\alpha \to \beta}_1...\mathcal{E}^{\alpha \to \beta}_N)$. The decoder takes $(\mathbf{D}^{\beta}_1...\mathbf{D}^{\beta}_N)$, $(\mathcal{E}^{\alpha \to \beta}_1...\mathcal{E}^{\alpha \to \beta}_N)$ as input and outputs $(\mathcal{D}^{\alpha \to \beta}_1...\mathcal{D}^{\alpha \to \beta}_N)$.
	
	\paragraph{Backward Translation}
	Backward Translation ensures the robustness of $\mathcal{E}^{\alpha \to \beta}_i$ by translating between modalities backwards (i.e. from modality $\beta$ to $\alpha$). The Backward Transformer is similar to the Forward Transformer, but the backward one translates from $(\mathcal{D}^{\alpha \to \beta}_1...\mathcal{D}^{\alpha \to \beta}_N)$ into $(\mathbf{X}^{\alpha}_1...\mathbf{X}^{\alpha}_N)$. The encoded features and the decoded features in Backward Transformer are denoted as $(\mathcal{E}^{\beta \to \alpha}_1...\mathcal{E}^{\beta \to \alpha}_N)$ and $(\mathcal{D}^{\beta \to \alpha}_1...\mathcal{D}^{\beta \to \alpha}_N)$ respectively.

	For each utterance $\mathbf{X}_i$, the Modality Fusion Cell produces four new features, $\mathcal{E}^{\alpha \to \beta}_i$, $\mathcal{E}^{\beta \to \alpha}_i$, $\mathcal{D}^{\alpha \to \beta}_i$, $\mathcal{D}^{\beta \to \alpha}_i$, from $\mathbf{X}^{\alpha}_i$ and $\mathbf{X}^{\beta}_i$. These features will be used later in this section to predict the sentiment tendency.
	\begin{figure}[t]
		\centering
		\includegraphics[width=1\linewidth]{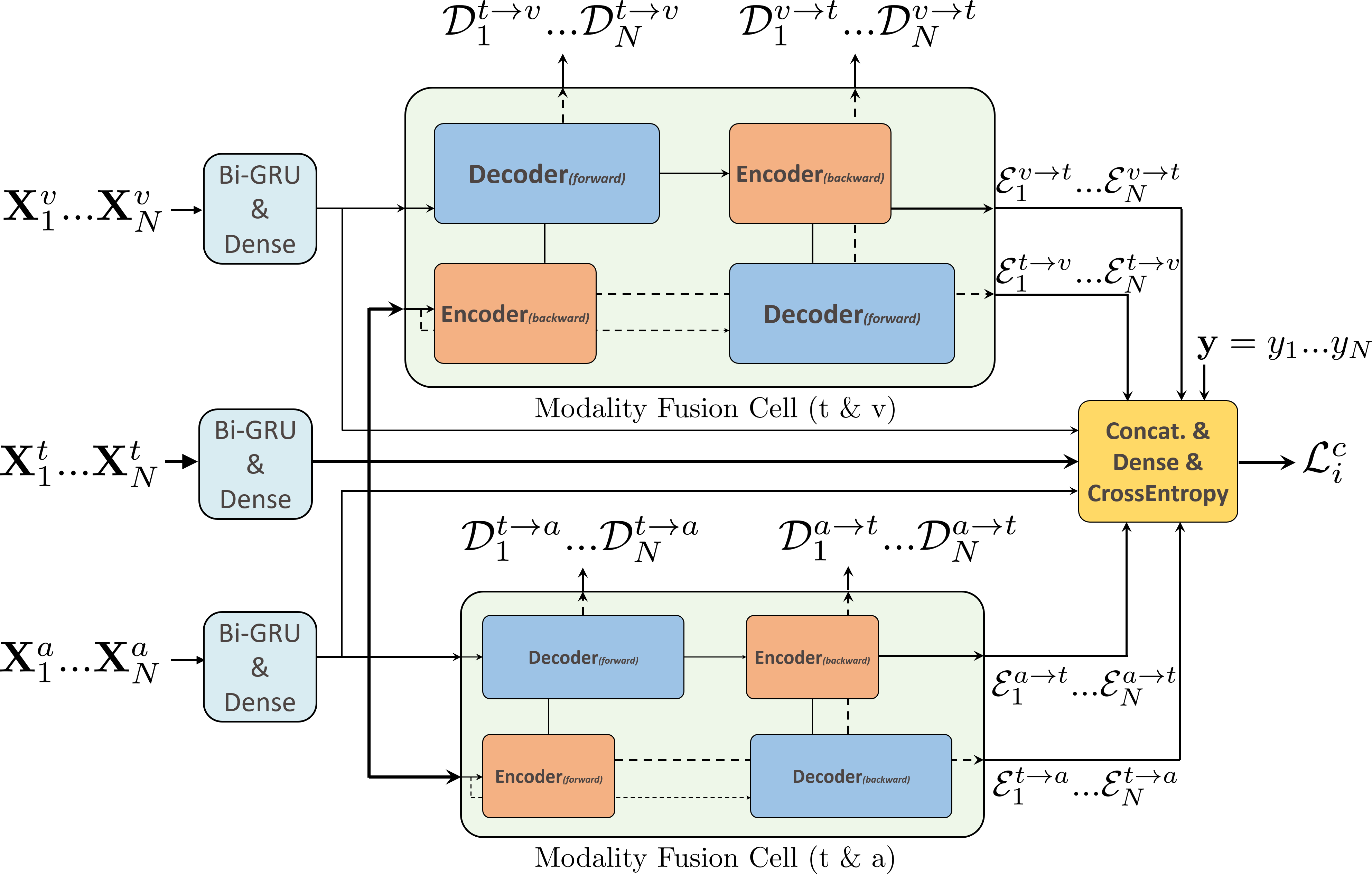}
		\caption{The architecture of TransModality}
		\label{fig:transmodality}
	\end{figure}

	\subsection{TransModality}
	Multimodal features are different in abilities to predict speaker's sentiment tendency. We adopt \textit{text} as the main modality, \textit{visual} and \textit{acoustic} as auxiliary modalities.
	
	We build our model, TransModality, with two Modality Fusion Cells for the \textit{text-visual} and \textit{text-acoustic} modality fusion, and a fully connected dense layer for classification. The two Modality Fusion Cells are built in parallel way, rather than sequential way, to eliminate the interference between modalities.
	
	\paragraph{Modality Fusion}
	We input (textual, visual) and (textual, acoustic) features into the two separate Modality Fusion Cells. Each Modality Fusion Cell enables the textual features to blend with the information from another modality. Since there are two Modality Fusion Cells in our model, we get four encoded features for utterance $\mathbf{X}_i$, $\mathcal{E}^{t \to v}_i$, $\mathcal{E}^{v \to t}_i$, $\mathcal{E}^{t \to a}_i$, $\mathcal{E}^{a \to t}_i$, which will be used for classification. The four decoded features for utterance $\mathbf{X}_i$, $\mathcal{D}^{t \to v}_i$, $\mathcal{D}^{v \to t}_i$, $\mathcal{D}^{t \to a}_i$, $\mathcal{D}^{a \to t}_i$, will be used to compare with corresponding target features to improve the translation performance.
	
	\paragraph{Translation Evaluation}
	The decoded features serves as the result of translation. Because the decoded features may be of different dimension from the target features, fully connected dense layers are used to project decoded features to the dimension of target features. Four separate fully connected dense layers are used and their outputs are denoted as $\hat{\mathbf{X}}^{\alpha \to \beta}_i$ where $(\alpha, \beta) \in \{(t,v),(v,t),(t,a),(a,t)\}$. The mean absolute error is used to train the translation model.
	\begin{equation}
	\mathcal{L}^{\alpha \to \beta}_i = \frac{1}{d_{\beta}} \sum_{j=1}^{d_{\beta}} \left| \hat{\mathbf{X}}^{\alpha \to \beta}_{ij} - \mathbf{X}^{\beta}_{ij} \right|
	\end{equation}
	where $d_{\beta}$ is the dimension of $\mathbf{X}^{\beta}_i$.
	
	For each utterance $\mathbf{X}_i$, we get four MAE losses to evaluate the translation performance: $\mathcal{L}^{t \to v}_i$, $\mathcal{L}^{v \to t}_i$, $\mathcal{L}^{t \to a}_i$, $\mathcal{L}^{a \to t}_i$.
	
	\paragraph{Classification Layer}
	The encoded features from the Modality Fusion Cells are considered as the joint features of the two modalities involved. Motivated by the residual skip connection network \cite{he2016deep}, all encoded features are concatenated to obtain the final joint features $\mathbf{F}_i$ which embodies information from all modalities available.
	\begin{equation}
	\mathbf{F}_i = [\mathcal{E}^{t \to v}_i, \mathcal{E}^{v \to t}_i, \mathcal{E}^{t \to a}_i, \mathcal{E}^{a \to t}_i, \mathbf{D}^t_i, \mathbf{D}^v_i, \mathbf{D}^a_i]
	\end{equation}
	
	Finally, $\mathbf{F}_i$ is passed to a fully connected dense layer to calculate the probability $\mathcal{P}_i$ and we use the cross entropy loss function.
	\begin{equation}
	\mathcal{P}_i = \mathrm{softmax}(W\mathbf{F}_i+b) \in \mathbb{R}^{d}
	\end{equation}
	\begin{equation}
	\hat{y}_i = \mathrm{argmax}_j(\mathcal{P}_i[j])
	\end{equation}
	\begin{equation}
	\mathcal{L}^{c}_i = -\log(\mathcal{P}_i[y_i])
	\end{equation}
	where $W$ is the weight matrix, $b$ is the bias. $d$ is the category number. $\mathcal{P}_i[j]$ refers to the possibility for label $j$ of utterance $\mathbf{X}_i$. $\hat{y}_i$ is the predicted label for utterance $\mathbf{X}_i$. $y_i$ is the true label for utterance $\mathbf{X}_i$.
	
	The joint loss of each utterance is a weighted summation of translation losses and classification loss ($\mathcal{L}^{t \to v}_i$, $\mathcal{L}^{v \to t}_i$, $\mathcal{L}^{t \to a}_i$, $\mathcal{L}^{a \to t}_i$, $\mathcal{L}^{c}_i$). The final loss of the whole model is the average of each utterance's joint loss.
	
	\subsection{Bi-TransModality}
	For a dataset with only two modalities, we propose Bi-TransModality, which is similar to TransModality but takes only two modality features as input, denoted as $\alpha$ and $\beta$. Only one Modality Fusion Cell is used. The final joint features are calculated as
	\begin{equation}
	\mathbf{F}_i = [\mathcal{E}^{\alpha \to \beta}_i, \mathcal{E}^{\beta \to \alpha}_i, \mathbf{D}^{\alpha}_i, \mathbf{D}^{\beta}_i]
	\end{equation}
	We also calculate the weighted joint loss to train the model.

	\section{Experiment Setting}
	
	\subsection{Datasets}
	We evaluate our proposed model on three multimodal datasets: CMU-MOSI, MELD and IEMOCAP. In each dataset, videos are separated into several utterance-level segments, and each utterance is annotated with a label showing its sentiment tendency or emotion tendency. Table \ref{tab:datadistribution} shows the distribution of train, validation and test samples in the datasets.
	\begin{table}[htbp]
		\centering
		\caption{Data distribution}
			\begin{tabular}{|c|c|c|c|c|}
				\hline
				Dataset & Modality & {Partition} & {Videos} & {Utterances} \\
				\hline
				\hline
				\multirow{2}[4]{*}{CMU-MOSI} & \multirow{2}[4]{*}{\textit{{t,v,a}}} & Train \& Valid & 62    & 1447 \\
				\cline{3-5}          &       & Test  & 31    & 752 \\
				\hline
				\multirow{2}[4]{*}{MELD} & \multirow{2}[4]{*}{\textit{{t,a}}} & Train \& Valid & 1152  & 11098 \\
				\cline{3-5}          &       & Test  & 280   & 2610 \\
				\hline
				\multirow{2}[4]{*}{IEMOCAP} & \multirow{2}[4]{*}{\textit{{t,v,a}}} & Train \& Valid & 120   & 5810 \\
				\cline{3-5}          &       & Test  & 31    & 1623 \\
				\hline
		\end{tabular}
		\label{tab:datadistribution}%
	\end{table}%
	\textit{CMU-MOSI \cite{zadeh2016mosi}} dataset contains opinion videos from online sharing websites such as YouTube. Each utterance is annotated as either positive or negative. We use utterance-level features provided in \citet{poria2017context} for fair comparison with MMMU-BA \cite{ghosal2018contextual}.
	
	\textit{MELD \cite{poria2018meld}} dataset is a new multimodal multiparty conversational dataset which collects actor's lines from \textit{Friends}, a famous American TV series. The dataset provides features of two modalities, textual and acoustic. Each utterance has been annotated with two labels. One shows its sentiment tendency among positive, neutral or negative. The other one shows its emotion tendency among anger, disgust, fear, joy, neutral, sadness and surprise (7 categories). We denote it as MELD (Sentiment) and MELD (Emotion). 
	
	\textit{IEMOCAP \cite{busso2008iemocap}} dataset contains conversation videos. Each video contains a single dialogue and is segmented into utterances. The utterances are annotated with one of 6 emotion labels: happy, sad, neutral, angry, excited and frustrated. 
	
	These features are pre-trained through CNN, 3D-CNN and openSMILE \cite{eyben2010opensmile} for textual, visual and acoustic features, respectively.

	\subsection{Baselines}
	To prove the effectiveness of our proposed method, we compare our TransModality model with the following strong baseline models: 
	
	\textit{bc-LSTM \cite{poria2017context}} use LSTM and dense layers to extract the contextual features and use simple concatenation as fusion method.

	\textit{CHFusion \cite{majumder2018multimodal}} (current state-of-the-art on IEMOCAP) fuses multimodal features through a hierarchical network. It first fuses each two modalities and then fuses all three modalities.

	\textit{MMMU-BA \cite{ghosal2018contextual}} (current state-of-the-art on CMU-MOSI) uses a complex attention mechanism as fusion method.

	\textit{GME-LSTM(A) \cite{chen2017multimodal}} uses Gated Embedding to alleviate the interference of noisy modalities, and the LSTM with Temporal Attention to fuse input modalities.

	\textit{Seq2Seq2Sent \cite{pham2018seq2seq2sentiment}} adopts hierarchical seq2seq model to get joint multimodal representation.

	\textit{MCTN \cite{pham2018found}} uses seq2seq model to fuse modalities. It builds a sequential fusion pipeline, which fuses textual and visual features at first, and then fuses the joint feature with acoustic features. 

	\textit{MELD-base \cite{poria2018meld}} is baseline in \citet{poria2018meld}. It extracts contextual information through GRU and fuses the multimodal features by concatenation.

	\section{Results and Discussion}
	We compare our proposed model with all baseline models in the multimodal datasets under different modality settings. Weighted accuracy score is used as evaluation metric. We also use sign test \cite{dixon1946statistical} to compare our model with the existing state-of-the-art models. The results are shown in Table \ref{tab:resulta}, \ref{tab:resultb}. All the experiments are conducted under the same settings, so the comparison is fair and trustworthy.

	Our proposed fusion method for bi-modal settings is Bi-TransModality, and TransModality is for tri-modal setting.  
	\begin{table*}[t]
		\centering
		\caption{Results on CMU-MOSI \& MELD (Sentiment) \\ * indicates \textit{p-value} < 0.05 for sign test when compared with our method}
		\begin{tabular}{|l|ccc|ccc|c|cc|c|}
		\hline
		\multirow{3}[6]{*}{} & \multicolumn{7}{c|}{CMU-MOSI}                         & \multicolumn{3}{c|}{MELD (Sentiment)} \\
		\cline{2-11}      & \multicolumn{3}{c|}{Uni} & \multicolumn{3}{c|}{Bi} & Tri   & \multicolumn{2}{c|}{Uni} & Bi \\
		\cline{2-11}      & \textit{{t}} & \textit{{v}} & \textit{{a}} & \textit{{v,a}} & \textit{{t,v}} & \textit{{t,a}} & \textit{{t,v,a}} & \textit{{t}} & \textit{{a}} & \textit{{t,a}} \\
		\hline
		\hline
		MELD-base & 77.79     & 55.19     & 55.85     & 54.79     & 76.60     & 76.99     & 79.19*     & 66.33  & 46.43  & 66.68* \\
		bc-LSTM & 79.12  & 55.98  & 57.31  & 56.52  & 78.59  & 78.86  & 79.26*  & 65.85  & 54.39  & 66.09*  \\
		CHFusion & -    & -    & -    & 54.49  & 74.77  & 78.54  & 76.51*  & -    & -    & 65.85*  \\
		MMMU-BA & -    & -    & -    & 57.45  & \textbf{80.85 } & 79.92  & 81.25*  & -    & -    & 65.56*  \\
		GME-LSTM(A) & 71.30     & 52.30     & 55.40     & 52.90  & 74.30  & 73.50  & 76.50*  & 65.52  & 52.03  & 66.46  \\
		seq2seq2sent & -    & -    & -    & 58.00  & 67.00  & 66.00  & 70.00*  & -    & -    & 63.84*  \\
		MCTN  & -    & -    & -    & 53.10  & 76.80  & 76.40  & 79.30*  & -    & -    & 66.27 \\
		\hline
		\textbf{Ours} & -    & -    & -    & \textbf{59.97 } & 80.58  & \textbf{81.25 } & \textbf{82.71 } & -    & -    & \textbf{67.04 } \\
		\hline
		\end{tabular}%
		\label{tab:resulta}%
	\end{table*}%

	\begin{table*}[t]
		\centering
		\caption{Results on IEMOCAP \& MELD (Emotion) \\ * indicates \textit{p-value} < 0.05 for sign test when compared with our method}
		\begin{tabular}{|l|cc|c|ccc|ccc|c|}
		\hline
		\multirow{3}[6]{*}{} & \multicolumn{3}{c|}{MELD (Emotion)} & \multicolumn{7}{c|}{IEMOCAP} \\
	\cline{2-11}          & \multicolumn{2}{c|}{Uni} & Bi    & \multicolumn{3}{c|}{Uni} & \multicolumn{3}{c|}{Bi} & Tri \\
	\cline{2-11}          & \textit{{t}} & \textit{{a}} & \textit{{t,a}} & \textit{{t}} & \textit{{v}} & \textit{{a}} & \textit{{v,a}} & \textit{{t,v}} & \textit{{t,a}} & \textit{{t,v,a}} \\
		\hline
		\hline
		MELD-base & 56.75  & 39.74  & 57.85*  & 55.51     & 39.06     & 47.57     & 48.31     & 56.62     & 57.12     & 58.29* \\
		bc-LSTM & 59.96  & 49.46  & 60.19  & 56.81  & 38.51  & 46.15  & 47.38  & \textbf{57.42 } & 57.55  & 58.23*  \\
		CHFusion & -    & -    & 58.35*  & -    & -    & -    & 41.84  & 56.83  & 57.30  & 58.90*  \\
		MMMU-BA & -    & -    & 60.26*  & -    & -    & -    & 49.66  & 57.30  & 57.18  & 58.78*  \\
		GME-LSTM(A) & 59.57  & 49.59  & 60.01*  & 56.69  & 39.86  & 48.98  & 48.55  & 56.44  & 56.93  & 57.98*  \\
		seq2seq2sent & -    & -    & 56.42*  & -    & -    & -    & 46.81  & 53.81  & 52.56  & 54.75*  \\
		MCTN  & -    & -    & 59.96*  & -    & -    & -    & 49.88  & 54.81  & 55.13  & 57.38*  \\
		\hline
		\textbf{Ours} & -    & -    & \textbf{61.95 } & -    & -    & -    & \textbf{50.15 } & 56.93  & \textbf{58.84 } & \textbf{60.81 } \\
		\hline	  
		\end{tabular}%
		\label{tab:resultb}%
	\end{table*}%
	\subsection{Comparison with Baselines}
	As evidenced by Table \ref{tab:resulta} and Table \ref{tab:resultb}, with more modalities involved, an improvement in prediction performance can be witnessed. In our proposed method, there is big accuracy improvement with all three modalities considered, compared to two-modality version. This suggests that the multimodality fusion can make extra contribution to the prediction performance.
	
	We can also observe that our model surpasses the existing state-of-the-art methods on the three datasets in most settings, including bi-modality settings and tri-modality settings. On CMU-MOSI dataset, our proposed method achieves about 1.6 points accuracy improvement. On MELD dataset, our proposed method achieves 0.33 points and 1.69 points accuracy improvement for sentiment and emotion analysis, respectively. On IEMOCAP dataset, our proposed method achieves about 1.4 accuracy improvement. The comparison on all three different datasets and even for different tasks (sentiment analysis and emotion analysis) well demonstrates TransModality can be widely adopted to better human sentiment and emotion analysis. We think the enhancement is caused by the new fusion methodology used by our model.
	
	We further perform sign test when comparing the performance of our method and baseline methods with all modalities, i.e., on the CMU-MOSi and IEMOCAP datasets, we perform sign test on the results with three modalities, and on the MELD dataset, we perform sign test on the results with two modalities. In Table \ref{tab:resulta} and Table \ref{tab:resultb}, we use * to indicate that the p-value for sign-test between the results of the baseline method and our method is less than 0.05, showing a statistically significant difference. We can see that our method can significantly outperform all the baseline methods on the CMU-MOSI dataset and the IEMOCAP dataset, and significantly outperform most baseline methods on the MELD dataset.
	
	Bc-LSTM, MMMU-BA and MELD-base utilize concatenation or attention mechanism to directly blend multimodal features, but features may interfere with each other when making predictions.
	
	GME-LSTM(A) and MCTN focus on the interference and refine the modality fusion, but information from different modalities cannot be blended well. The LSTM with Temporal Attention in GME-LSTM(A) is limited in the ability to fuse multimodal features together. The sequential fusion pipeline in MCTN also increases the risk of interference between modalities.
	
	In our proposed model TransModality, fusion between modalities is conducted by end-to-end translation with Transformer. The encoded features serve as the joint features of the two modalities involved. Our joint features are encoded from the source modality features by Multi-Head Attention and Feed Forward Layers, so they contain most information of the source modality. The target modality features can be decoded from the joint features as the result of Transformer, so we assume that the information from target modality has also been well blended into our joint features.
	
	Compared with the baselines, our model conducts no direct calculation on the multimodal features and the pair-wise fusion is conducted in parallel way, so it can better prevent the interference. Our method also ensures the modality fusion performance through Forward and Backward Translation. 
	
	Our new idea on the fusion methodology improves the performance and contributes to a better prediction result, and the effectiveness of our proposed method under different circumstances has been proven clearly from the comparison.
	
	\subsection{Analysis of Translation}
	The end-to-end translation is the core idea of our model. We select Transformer as the major component, use parallel translation to eliminate modality interference, and adopt Forward and Backward Translation to improve the performance. 
	
	We conduct head-to-head experiment to compare with \citet{pham2018seq2seq2sentiment,pham2018found} to show the contribution of Transformer and other components in our model. Then we also demonstrate the effectiveness of Forward and Backward Translation through comparative experiments between different versions of our model.
	
	\paragraph{Comparison with Seq2seq Translation}
	We replace the seq2seq in MCTN \cite{pham2018found} with Transformer, and build a model called Trans-MCTN. This model is the same as MCTN except for the use of Transformer. We compare the result of seq2seq2sent, MCTN, Trans-MCTN and TransModality on all three datasets to show the improvement from Transformer. The results are listed in Table \ref{tab:head2head}.
	\begin{table}[htbp]
		\centering
		\caption{Results of Seq2seq-based or Transformer-based Translation \\ $^{\blacktriangle}$ indicates \textit{p-value} < 0.05 (compared with Trans-MCTN) \\ * indicates \textit{p-value} < 0.05 (compared with TransModailty)}
		\begin{tabular}{|l|c|c|}
		\hline
		\multirow{2}[4]{*}{} & CMU-MOSI & MELD(Sentiment) \\
	\cline{2-3}          & t,v,a & t,a \\
		\hline
		\hline
		seq2seq2sent & 70.00*$^{\blacktriangle}$ & 63.84*$^{\blacktriangle}$ \\
		MCTN  & 79.30*$^{\blacktriangle}$ & 66.27$^{\blacktriangle}$ \\
		\hline
		Trans-MCTN & 81.67 & 66.16* \\
		\hline
		\textbf{TransModality} & \textbf{82.71} & \textbf{67.04} \\
		\hline
		\multicolumn{1}{l}{} & \multicolumn{1}{c}{} & \multicolumn{1}{c}{} \\
		\hline
		\multirow{2}[4]{*}{} & MELD(Emotion) & IEMOCAP \\
	\cline{2-3}          & t,a   & t,v,a \\
		\hline
		\hline
		seq2seq2sent & 56.42*$^{\blacktriangle}$ & 54.75*$^{\blacktriangle}$ \\
		MCTN  & 59.96* & 57.38*$^{\blacktriangle}$ \\
		\hline
		Trans-MCTN & 60.24* & 58.72* \\
		\hline
		\textbf{TransModality} & \textbf{61.95} & \textbf{60.81} \\
		\hline
		\end{tabular}%
		\label{tab:head2head}%
	\end{table}%
	The models in the first two lines use seq2seq to translate. The models in the next two lines use Transformer to translate. With the p-values from sign test, the stable increment is observed in the comparison. On all three datasets, Trans-MCTN achieves an improvement compared to MCTN, and with other innovations in our proposed method, TransModality achieves an improvement compared to Trans-MCTN. 
	
	The experiment demonstrates that Transformer contributes to better performance than simple seq2seq method. Other innovations in our proposed method, such as the parallel translation, also refine the architecture to predict sentiment tendency more accurately. 

	\paragraph{Backward Translation}
	We further analyze our proposed architecture \textit{with} and \textit{without} Backward Translation. In the architecture for comparison, only one-way translation (i.e. only Forward Translation) is conducted.
	\begin{table}[t]
		\centering
		\caption{Results of using or not using Backward Translation}
		\begin{tabular}{|c|cc|cc|}
			\hline
			\multirow{2}[4]{*}{Modal} & \multicolumn{2}{c|}{CMU-MOSI} & \multicolumn{2}{c|}{MELD (Sentiment)} \\
		\cline{2-5}          & with  & without & with  & without \\
			\hline
			\hline
			\textit{{v, a}} & \textbf{59.97} & 58.78 & -     & - \\
			\textit{{t, v}} & \textbf{80.58} & 79.12 & -     & - \\
			\textit{{t, a}} & \textbf{81.25} & 79.78 & \textbf{67.04} & 66.02 \\
			\hline
			\textit{{t, v, a}} & \textbf{82.71} & 80.18 & -     & - \\
			\hline
			\multicolumn{1}{r}{} &       & \multicolumn{1}{r}{} &       & \multicolumn{1}{r}{} \\
			\hline
			\multirow{2}[4]{*}{Modal} & \multicolumn{2}{c|}{MELD (Emotion)} & \multicolumn{2}{c|}{IEMOCAP} \\
		\cline{2-5}          & with  & without & with  & without \\
			\hline
			\hline
			\textit{{v, a}} & -     & -     & \textbf{50.15} & 48.80 \\
			\textit{{t, v}} & -     & -     & \textbf{56.93} & 56.07 \\
			\textit{{t, a}} & \textbf{61.95} & 60.58 & \textbf{58.84} & 57.98 \\
			\hline
			\textit{{t, v, a}} & -     & -     & \textbf{60.81} & 59.21 \\
			\hline
			\end{tabular}%
		\label{tab:compareab}%
	\end{table}%
	From Table \ref{tab:compareab}, we observe that using Backward Translation indeed helps improve the performance of both TransModality and Bi-TransModality. There is approximately 1.5 points accuracy improvement on each dataset in different settings. 

	The comparative experiments suggests that Backward Translation assists Forward Translation and improves the performance. Forward and Backward Translation is important components in our proposed model.	After the careful analysis of translation performance and Backward Translation, our hypothesis has been verified that TransModality is a reliable method to conduct modality fusion.
	
	\section{Conclusion}
	In this paper, we propose a new end-to-end translation based multimodal fusion method for utterance-level sentiment analysis. Our method utilizes Transformer to translate between modalities and blends multimodal information into the encoded features. With Transformer, the encoded features can embody the information from both the source modality and the target modality. We also adopt Forward and Backward Translations to better model the correlation between multimodal features and improve the translation performance. Through the validation results on multiple multimodal datasets (CMU-MOSI, MELD, IEMOCAP), we demonstration the reliability of Transformer in multimodality fusion and the effectiveness of Forward and Backward Translations.
	
	In future work, we will apply our model to other multimodal classification tasks, such as multimodal sarcasm detection and stance detection, to further test the robustness of the model. 

	\section{Acknowledgments}
	This work was supported by National Natural Science Foundation of China (61772036) and Key Laboratory of Science, Technology and Standard in Press Industry (Key Laboratory of Intelligent Press Media Technology). We appreciate the anonymous reviewers for their helpful comments. Xiaojun Wan is the corresponding author.

\bibliographystyle{ACM-Reference-Format}
\bibliography{ref}

\end{document}